\documentclass[nohyperref]{article}

\usepackage{microtype}
\usepackage{graphicx}
\usepackage{booktabs} %

\usepackage{hyperref}

\usepackage[accepted]{icml2023}

\usepackage{amsmath}
\usepackage{amssymb}
\usepackage{mathtools}
\usepackage{amsthm}

\usepackage[capitalize,noabbrev]{cleveref}

\theoremstyle{plain}

\theoremstyle{definition}

\theoremstyle{remark}

\usepackage{pifont}%
\usepackage{xcolor}%
\usepackage{mathtools}%
\usepackage{subcaption}
\usepackage{caption}
\usepackage{adjustbox}
\usepackage{arydshln}
\usepackage{tabularx}
\usepackage{makecell}
\usepackage{lipsum}

\mathtoolsset{showonlyrefs=true} %

\newcommand{\sdval}[2]{$#1 \pm #2$}

\newcommand{\hp}[1]{\hphantom{#1}}

\newcommand{\timeschange}[2]{$\mathit{#1\times - #2\times}$}

\newcommand{\atol}{\texttt{atol}}
\newcommand{\rtol}{\texttt{rtol}}
\newcommand{\eest}{\texttt{E}_{\texttt{Est}}}
\newcommand{\treg}{\texttt{t}_{\texttt{reg}}}

\newcommand{\func}[2]{#1\left(#2\right)}

\newcommand{\cpaper}{~$^\text{\textparagraph}$}

\newcommand{\bigO}[1]{\func{\mathcal{O}}{#1}}

\icmltitlerunning{Locally Regularized Neural Differential Equations}

\begin{document}

\twocolumn[
\icmltitle{Locally Regularized Neural Differential Equations:\\ Some Black Boxes Were Meant to Remain Closed!}

\icmlsetsymbol{equal}{*}

\begin{icmlauthorlist}
\icmlauthor{Avik Pal}{csail,miteecs}
\icmlauthor{Alan Edelman}{csail,mitmath}
\icmlauthor{Christopher Rackauckas}{csail}
\end{icmlauthorlist}

\icmlaffiliation{csail}{CSAIL MIT}
\icmlaffiliation{mitmath}{Department of Mathematics MIT}
\icmlaffiliation{miteecs}{Department of Electrical Engineering and Computer Science MIT}

\icmlcorrespondingauthor{Avik Pal}{avikpal@mit.edu}

\icmlkeywords{Machine Learning, Implicit Neural Networks, Neural ODEs, ICML}

\vskip 0.3in
]

\printAffiliationsAndNotice{}  %

\begin{abstract}

    Implicit layer deep learning techniques, like Neural Differential Equations, have become an important modeling framework due to their ability to adapt to new problems automatically. Training a neural differential equation is effectively a search over a space of plausible dynamical systems. However, controlling the computational cost for these models is difficult since it relies on the number of steps the adaptive solver takes. Most prior works have used higher-order methods to reduce prediction timings while greatly increasing training time or reducing both training and prediction timings by relying on specific training algorithms, which are harder to use as a drop-in replacement due to strict requirements on automatic differentiation. In this manuscript, \textit{we use internal cost heuristics of adaptive differential equation solvers at stochastic time-points to guide the training towards learning a dynamical system that is easier to integrate}. We ``close the blackbox'' and allow the use of our method with any adjoint technique for gradient calculations of the differential equation solution. We perform experimental studies to compare our method to global regularization to show that we attain similar performance numbers without compromising on the flexibility of implementation on ordinary differential equations (ODEs) and stochastic differential equations (SDEs). \textit{We develop two sampling strategies to trade-off between performance and training time}. Our method reduces the number of function evaluations to \timeschange{0.556}{0.733} and accelerates predictions by \timeschange{1.3}{2}.

\end{abstract}

\begin{table*}[t]
  \centering
  \adjustbox{max width=\textwidth}{
    \centering
    \begin{tabular}{lcc}
      \toprule
      \thead{Sensitivity Algorithm}                                          & \thead{Memory Requirement}                 & \thead{Memory Requirement with Local Regularization}         \\
      \midrule
      Backsolve Adjoint \citep{chen2018neural}                               & $\bigO{s}$                                 & $\bigO{s \times (1 + \mathrm{stages})}$                                \\
      Backsolve Adjoint with Checkpointing \citep{chen2018neural}            & $\bigO{s \times c}$                        & $\bigO{s \times (c  + \mathrm{stages})}$                      \\
      Interpolating Adjoint \citep{hindmarsh2005sundials}                    & $\bigO{s \times t}$                        & $\bigO{s \times (t  + \mathrm{stages})}$                      \\
      Interpolating Adjoint with Checkpointing \citep{hindmarsh2005sundials} & $\bigO{s \times c}$                        & $\bigO{s \times (c + \times \mathrm{stages})}$                       \\
      Quadrature Adjoint \citep{kim2021stiff}                                & $\bigO{\left(s + p\right) \times t}$       & $\bigO{\left(s + p\right) \times t + s \times \mathrm{stages}}$      \\
      Direct Reverse Mode Differentiation                                    & $\bigO{s \times t \times \mathrm{stages}}$ & $\bigO{s \times \left(t + 1\right) \times \mathrm{stages}}$ \\
      \bottomrule
    \end{tabular}
  }
  \caption{\textbf{Memory Requirements for various Sensitivity Algorithms for ODEs with Local Regularization}: Local Regularization requires $\bigO{s \times (c  + \mathrm{stages})}$ memory in the best case, while Global Regularization~\citep{pal2021opening} always requires $\bigO{s \times t \times \mathrm{stages}}$ memory.}
  \label{tab:memory_requirements_sensitivity_analysis_odes_with_local_reg}
\end{table*}

\begin{figure*}[t]
    \centering
    \includegraphics[width=\textwidth]{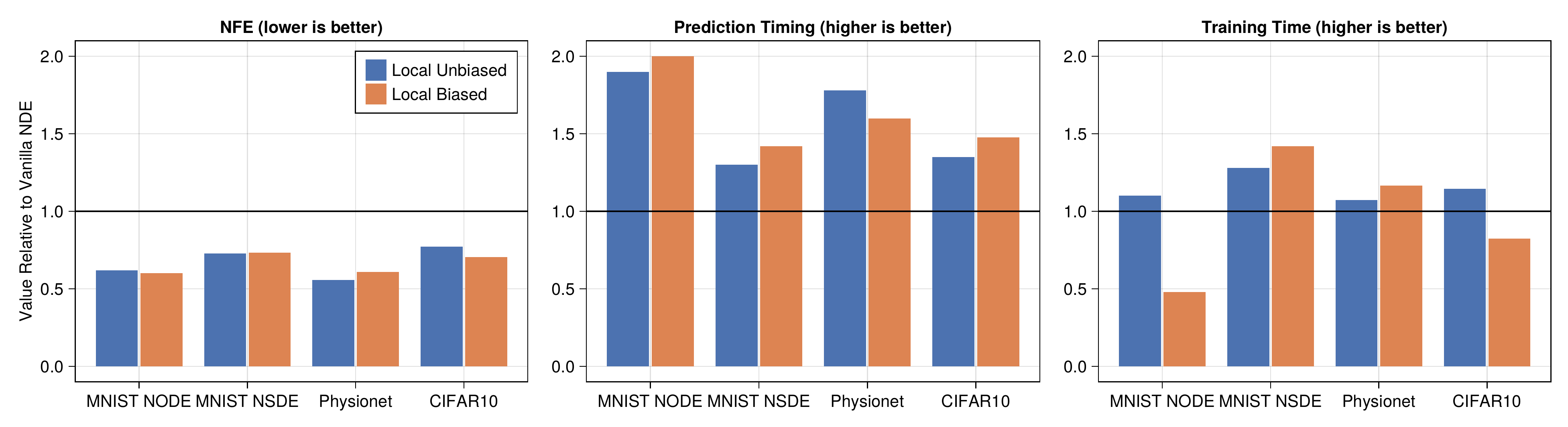}
    \caption{\textbf{Locally Regularized NDE leads to faster predictions and faster training compared to vanilla NDE.}}
    \label{fig:summary_plot}
\end{figure*}

\section{Introduction}
\label{sec:introduction}

Implicit Models, such as Neural Ordinary Differential Equations~\cite{chen2018neural} and Deep Equilibrium Models~\cite{bai2019deep, pal2022mixing}, have emerged as a promising technique to determine the depth of neural networks automatically. To maximize performance on a dataset, explicit models are tuned to the ``hardest'' training sample, which hurts the inference timings for ``easier'' -- more abundant -- samples. Using adaptive differential equation solvers allow these implicit models to choose the number of steps they need effectively. This idea of representing neural networks as ODEs has since been generalized to Stochastic Differential Equations~\citep{liu2019neural, rackauckas2020universal} and other architectures to improve robustness.

Despite the rapid progress in these methods, the core problem of the scalability of these models is still persistent. Several solutions to these have been proposed:
\begin{itemize}
    \item \citet{kelly2020learning, finlay2020train} use higher order derivatives for regularization.
    \item \citet{poli2020hypersolvers} learn neural solvers to solve Neural ODEs faster.
    \item \citet{pal2021opening} proposed a ``zero-cost'' global regularization scheme. 
    \item \citet{ghosh2020steer} randomize the integration stop time to ``smoothen'' the dynamics. 
\end{itemize}
All these methods have definite tradeoffs (See \Cref{sec:related_works} for more details).

This paper presents a generally applicable method to force the neural differential equation training process to choose the least expensive option. We build upon the global regularization scheme proposed in \citet{pal2021opening} and ``close'' the blackbox allowing our method to work across various sensitivity algorithms. Our main contributions include the following\footnote{Our code is publicly available at \url{https://github.com/avik-pal/LocalRegNeuralDE.jl}}:
\begin{enumerate}
    \item We show that our local regularization method -- building upon the primitives proposed in \citet{pal2021opening} -- performs at par with global regularization (See \Cref{sec:experiments}).

    \item We present two sampling methods that trade-off small computational costs for consistently better performance (See \Cref{sec:methods}).

    \item Using local regularization allows our models to leverage optimize-then-discretize in the backward pass (in addition to discretize-then-optimize methods). Our method works around the several engineering limitations of automatic differentiation (AD) systems~\citep{rackauckas_2022} that are needed to make global regularization work.

    \item We empirically show that regularizing solver heuristics with biased sampling stabilizes the training of larger neural ODEs (See \Cref{subsec:cifar10}).
\end{enumerate}

\section{Background}
\label{sec:background}

\subsection{Neural Ordinary Differential Equations}
\label{subsec:neural_odes}

Neural ODEs use an explicit neural network to parameterize the dynamical system. These involve finding the state at a later time $t_1$, given the value $z_0$ at time $t_0$. The final state:
\begin{equation}
    z(t_1) = z_0 + \int_{t_0}^{t_1} \func{f_\theta}{z(t), t} dt
\end{equation}
generally cannot be computed analytically and requires numerical solvers. Prior works had observed the similarity between fixed time-step discretization of ODEs and Residual Neural Networks~\citet{lu2018beyond}, which was later extended in the Neural ODE framework by \citet{chen2018neural}.
\begin{equation}
    \frac{dz(t)}{dt} = \func{f_\theta}{z(t), t}
\end{equation}
Using adaptive time stepping allows the model to operate at a variable continuous depth depending on the inputs. Removal of the fixed depth constraint of Residual Networks provides a more expressive framework and offers several advantages in problems like density estimation~\citep{grathwohl2018ffjord}, irregularly spaced time series problems~\citep{rubanova2019latent}, etc.

\subsection{Neural Stochastic Differential Equations}
\label{subsec:neural_sdes}

Stochastic Differential Equations (SDEs) couple the effect of noise to a deterministic system of equations. SDE noise can have different variations; however, we exclusively focus on diagonal multiplicative noise for this paper. \citet{liu2019neural} modified Neural ODEs by stochastic noise injection in the form of Neural SDEs. \citet{liu2019neural} empirically showed that stochastic noise injection improves the robustness and generalization performance of neural ODEs. In Neural SDEs, we simultaneously train two explicit neural networks -- drift $f_\theta(z(t), t)$ and diffusion $g_\phi(z(t), t))$ s.t.:
\begin{equation}
    dz(t) = f_\theta(z(t), t) dt + g_\phi(z(t), t)) dW
\end{equation}

\subsection{Adaptive Time Stepping for Differential Equations}
\label{subsec:adaptive_time_stepping}

Runge-Kutta (RK) Methods~\citep{runge1895numerische, kutta1901beitrag} are widely used to approximate the solutions of ordinary differential equations numerically. Given a tableau of coefficients $\left\{A, c, b\right\}$, these methods combine $s$ stages to obtain the estimate at $t + dt$.
\begin{align}
    & k_s = f\left(t + c_s \cdot dt, z(t) + \sum_{i = 1}^{s - 1} a_{si} \cdot k_i\right)\\
    & z(t + dt) = z(t) + dt \cdot \left( \sum_{i = 1}^s b_i \cdot k_i \right)
\end{align}
Adaptive solvers need to maximize the step size ($dt$) while keeping the error estimate below the user-specified tolerances, i.e., they need to satisfy:
\begin{equation}
    \eest \leq \atol + \texttt{max}\left( |z(t)|, |z(t + dt)|\right) \cdot \rtol
\end{equation}
where $\eest$ is the local error estimate. Adaptive solvers utilize an additional linear combiner $\Tilde{b}_i$ to get an alternate solution, typically with one order less convergence~\citep{wanner1996solving, fehlberg1968classical, dormand1980family, Tsit5}.
\begin{equation}
    \Tilde{z}(t + dt) = z(t) + dt \cdot \left( \sum_{i = 1}^s \Tilde{b}_i \cdot k_i \right)
\end{equation}
A classic result from Richardson extrapolation shows that $\texttt{E}_{\texttt{Est}} = \| \Tilde{z}(t + dt) - z(t + dt) \|$ is an estimate of the local truncation error~\citep{hairer1, ascher1998computer}. The new step size is determined using the following:
\begin{equation}
    q = \left\| \frac{\eest}{\atol + \texttt{max}\left( |z(t)|, |z(t + dt)|\right) \cdot \rtol} \right\|
\end{equation}
\begin{itemize}
    \item If $q < 1$, $dt$ is accepted.
    \item Otherwise, it is rejected and reduced. A standard PI controller proposes the new step size to be:
    \begin{equation}
        dt_{new} = \eta \cdot q_{n - 1}^\alpha \cdot q_{n}^\beta \cdot dt
    \end{equation}
    where $\eta$ is the safety factor, $q_{n - 1}$ is the error proportion of the previous step, and $(\alpha, \beta)$ are tunable PI-gain hyperparameters~\cite{wanner1996solving}. 
\end{itemize}

We defer the discussion of error estimation schemes for stochastic RK integrators to  \citet{rackauckas2017adaptive, rackauckas2020sosri}.

\begin{figure*}
    \makeatletter
    \def\@captype{algocf}
    \makeatother
    \begin{minipage}{0.48\textwidth}
        \begin{algorithm}[H]
            \textbf{Data:} $\texttt{x}$, $f_\theta$, $t_{span}$\;

            \textbf{Result:} $\texttt{sol}$, $\texttt{r}$\;

            Define $\frac{du}{dt} = f_\theta(u, t)$\;

            $t_0, t_1 \gets t_{span}$\;

            $\treg \sim \mathbb{U}\left[t_0, t_1\right]$\;

            $\texttt{sol} \gets \texttt{solve}(\frac{du}{dt}, \texttt{ DE Solver},~t_{span})$\;

            $u_{\treg} \gets \texttt{sol}(\treg)$\;

            Run single step for the solver with time-span $(\treg, t_1)$\;

            $\texttt{r} \gets $ Local Error Estimate $@ t=\treg$\;
            \caption{\textbf{Unbiased Sampling}: Training Phase}
            \label{alg:unbiased_sampling}
        \end{algorithm}
    \end{minipage}%
    \hfill
    \begin{minipage}{0.48\textwidth}
        \begin{algorithm}[H]
            \textbf{Data:} $\texttt{x}$, $f_\theta$, $t_{span}$\;

            \textbf{Result:} $\texttt{sol}$, $\texttt{r}$\;

            Define $\frac{du}{dt} = f_\theta(u, t)$\;

            $t_0, t_1 \gets t_{span}$\;

            $\texttt{sol} \gets \texttt{solve}(\frac{du}{dt}, \texttt{ DE Solver},~t_{span})$\;

            $\treg \sim \mathbb{U}\left(\texttt{sol}.t\right)$\;

            $u_{\treg} \gets \texttt{sol}(\treg)$\;

            Run single step for the solver with time-span $(\treg, t_1)$\;

            $\texttt{r} \gets $ Local Error Estimate $@ t=\treg$\;
            \caption{\textbf{Biased Sampling}: Training Phase}
            \label{alg:biased_sampling}
        \end{algorithm}
    \end{minipage}
\end{figure*}

\subsection{Global Regularization using Local Error Estimates}
\label{subsec:global_regularization}

In \Cref{subsec:adaptive_time_stepping}, we observe how larger local error estimates $\texttt{E}_{\texttt{Est}}$ contribute to reduced step sizes and, in turn, higher training and prediction times for Neural Differential Equations. \citet{pal2021opening} proposed a regularization scheme to minimize the ``total local error in order to learn Neural ODEs with as large step sizes as possible.'' They compute the regularization term $\left(\mathcal{R}_{E}\right)_g$ as:
\begin{equation}
    \left(\mathcal{R}_{E}\right)_g = \sum_j \left(\texttt{E}_{\texttt{Est}}\right)_j \cdot |dt_j|
\end{equation}
where the summation is over the $j$ time steps of the solution. All the local error estimates $\texttt{E}_{\texttt{Est}}$ are precomputed by adaptive differential equation solvers, which makes the forward pass effectively have zero additional overhead.

Continuous adjoint methods~\citep{chen2018neural} define the derivatives in terms of the ODE quantities. The terms $\left(\texttt{E}_{\texttt{Est}}\right)_j = \sum_{i = 1}^s (b_i - \Tilde{b}_i) \cdot k_i$ cannot be constructed directly from the trajectory of the ODE solution, since $k_i$ terms are defined by the solver method (and not by the continuous ODE solution). As a result \citet{pal2021opening} performs discrete sensitivity analysis to compute the derivatives w.r.t. the regularization terms.

Discrete Sensitivity analysis is more stable than continuous adjoints~\citep{zhang2014fatode} and stabilizes the training process of Neural ODEs~\citep{onken2020discretize}. However, these methods are more memory intensive than continuous adjoints. Another shortcoming is the advanced tooling mechanism needed to differentiate through differential equation solvers~\citep{rackauckas_2022}, making this method hard to adopt. Since adaptive ODE solvers change the number of steps based on the values in the differential equation, discrete sensitivity analysis requires dynamic AD tooling which is generally less optimized than those built for static graph cases. We note that there are checkpointing methods to reduce the memory overhead~\citep{dauvergne2006data}, though these generally have focused on static graph AD systems. 

\subsection{Multiscale Neural ODE with Input Injection}
\label{subsec:multiscale_neural_ode}

Multiscale modeling~\citep{burt1987laplacian} has been the central theme for several deep computer vision applications~\citep{farabet2012learning, yu2015multi, chen2016attention, chen2017deeplab}. Standard Neural ODEs set the initial condition as input $x$ from the previous layer.  Multiscale Neural ODE with Input Injection is based on the Multiscale Deep Equilibrium Models (DEQs)~\citep{bai2019deep, bai_multiscale_2020, pal2022mixing}. If our model operates at $n$ scales, the initial condition $\{(z_0)_i\}_{i \in [n - 1]}$ is defined as $(z_0)_0 = x$ and $(z_0)_{i \in \{1, \dots n - 1\}} = 0$. Every feature set is upsampled or downsampled and fed into the other scales using a neural network. In line with Multiscale DEQs, we also repeatedly inject the original input $x$ at every step of the dynamical system.

\section{Methods}
\label{sec:methods}

In \Cref{subsec:global_regularization}, we discussed the downsides of using global regularization with local error estimates. To summarize:
\begin{enumerate}
    \item Global Regularization relies on discrete sensitivity analysis, which is \textit{more memory intensive}.

    \item Global Regularization depends on AD tooling to support dynamic compute graphs in an efficient way, making it \textit{hard to incorporate into existing codebases}.
\end{enumerate}
To get around these limitations, we developed a new technique using local sampling of error estimates at specific time points, rather than globally over the full interval.
We deal with sampling the ``appropriate'' time point for regularization by two strategies:
\begin{itemize}
    \item \textbf{\Cref{alg:unbiased_sampling} Unbiased Sampling}: We random uniformly sample the time-point in the integration time span. Intuitively, since we will perform the training for ``a large number of steps,'' the learned dynamical system would end up being faster to solve ``everywhere'' over the time span.

    \item \textbf{\Cref{alg:biased_sampling} Biased Sampling}: Adaptive Time-Stepping Differential Equation Solvers naturally take more steps around the area, which is harder to integrate. We can bias the regularization to operate around parts of the dynamical system which are ``harder'' by sampling a time-point from the solution time points.
\end{itemize}

\subsection{Unbiased Sampling}
\label{subsec:unbiased_sampling}

When training a Neural ODE, the integration timespan is fixed. Training any deep learning model involves several thousand steps. We compute the total local error estimate over the entire timespan when performing global regularization. For unbiased sampling, we hypothesize that if we regularize at random uniformly sampled time points in the timespan, the learned dynamical system will demonstrate similar properties in terms of NFE compared to global regularization. Our new regularization term becomes $\left(\mathcal{R}_{E}\right)_{\texttt{unbiased}}$ as:
\begin{equation}
    \left(\mathcal{R}_{E}\right)_{\texttt{unbiased}} = \left(\texttt{E}_{\texttt{Est}}\right)_{\treg} \cdot |dt_{\treg}| \quad \treg \sim \mathbb{U}[\texttt{tspan}]
\end{equation}

\subsection{Biased Sampling}
\label{subsec:biased_sampling}

\begin{figure}[t]
    \centering
    \includegraphics[width=0.9\linewidth]{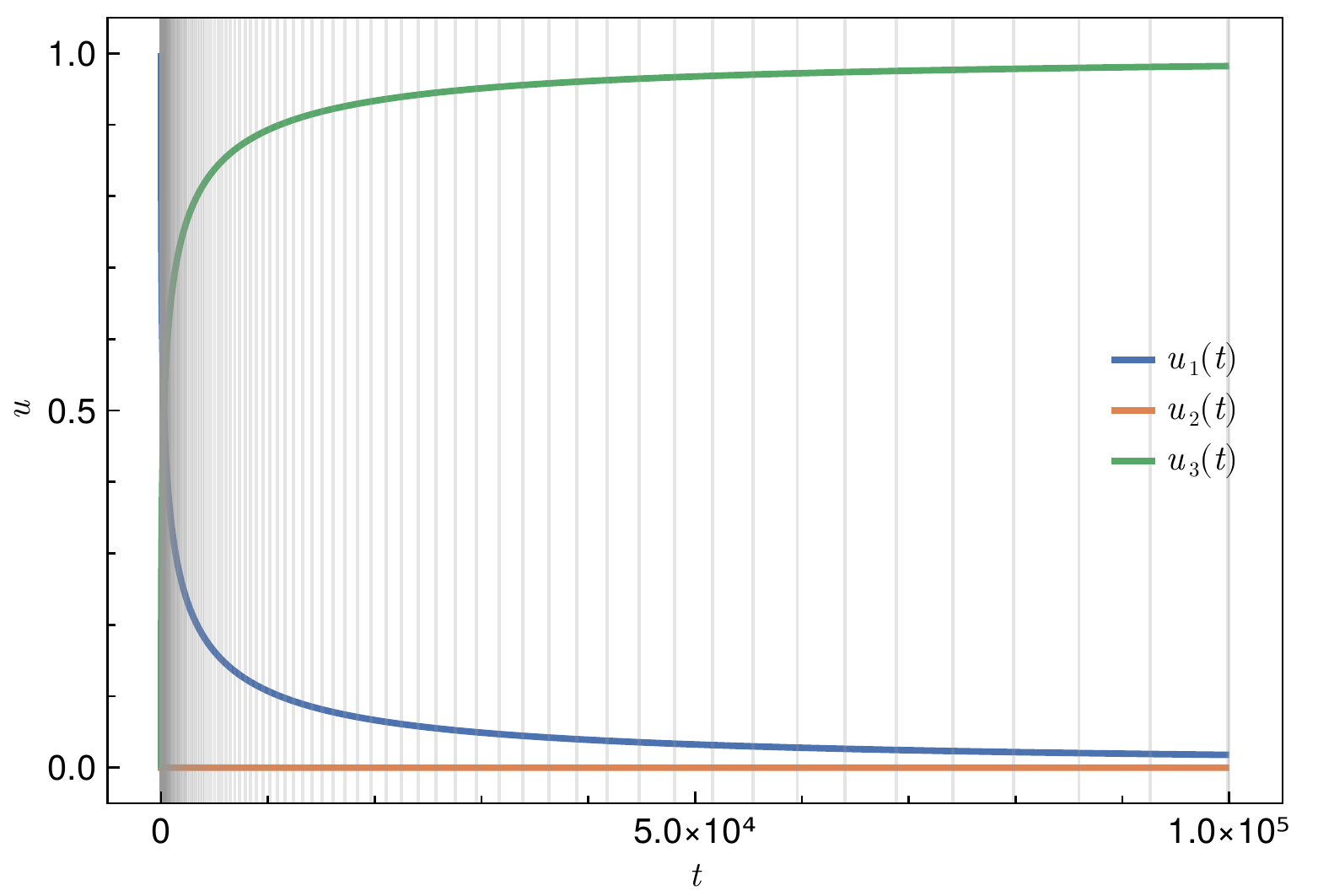}
    \caption{\textbf{Robertson Stiff ODE System}: Solving stiff systems like Robertson~\cite{robertson1966solution} (using Rodas5~\cite{piche1995stable}) involves spending around $75\%$ of the time in $t < 5000$ (i.e. $5\%$ of the timespan). The vertical lines denote the timepoints at which the ODE System was solved.}
    \label{fig:robertson_stiff_system}
    \vspace{-1.5em}
\end{figure}

Consider a simple scenario where the learned dynamics of the DE is harder to solve in $\left[0.25, 0.35\right]$, and we are solving the DE from $t_0 = 0$ to $t_1 = 1$. Our primary aim is to modify the learned system s.t. it becomes simpler to solve in $\left[0.25, 0.35\right]$. If we use unbiased sampling, the probability that we regularize at $\treg \in \left[0.25, 0.35\right]$ is $0.1$ (which is low). The problem gets even more severe if the range is lowered. An extreme version of this problem is observed for stiff systems like Robertson's Equations~(See \Cref{fig:robertson_stiff_system}) where $75\%$ of the time is spent in solving $5\%$ of the problem. We note that these extreme scenarios rarely occur for traditional deep learning tasks since \citet{pal2021opening} observed minor speedups using stiffness regularization. However, the problem that some parts of the dynamical system are harder to integrate persists, and designing a regularization scheme targeting those parts is highly desirable.

We considered a simple scenario where the learned dynamical system was fixed. However, while training NDEs, this system evolves with training, and apriori predicting the more difficult portions to integrate is not feasible. Adaptive solvers take more frequent steps in the parts of the DE where it is harder to integrate. \citet{anantharaman2020accelerating} leveraged this property of adaptive solvers to learn surrogates for stiff systems. Since these solvers adapt to concentrate around the most numerically difficult time points, we automatically obtain the time points where we want to regularize the model. Hence, for our biased sampling regularize, we uniformly sample the regularization timepoint $\treg$ from the time points at which the solver solved the differential equation.

\begin{table*}[t]
    \centering
    \adjustbox{max width=0.9\textwidth}{
        \centering
        \begin{tabular}{llllll}
            \toprule
            \thead{Method} & \thead{Train Accuracy (\%)} & \thead{Test Accuracy (\%)} & \thead{Training Time (hr)} & \thead{Prediction Time\\ (s / batch)} & \thead{Testing NFE}\\
            \midrule
            Vanilla NODE          & \sdval{99.898}{0.066} & \sdval{97.612}{0.163} & \sdval{0.54}{0.001} & \sdval{0.088}{0.020} & \sdval{303.559}{3.194}\\
            STEER\cpaper          & \sdval{100.00}{0.000} & \sdval{97.94\hp{0}}{0.03\hp{0}} & \sdval{1.31}{0.07\hp{0}} & \sdval{0.092}{0.002} & \sdval{265.0\hp{0}\hp{0}}{3.46\hp{0}}\\
            TayNODE\cpaper        & \sdval{\hp{0}98.98}{0.06} & \sdval{97.89\hp{0}}{0.00} & \sdval{1.19}{0.07} & \sdval{0.079}{0.007} & \sdval{\hp{0}80.3\hp{0}\hp{0}}{0.43}\\
            ERNODE\cpaper         & \sdval{\hp{0}99.71}{0.28} & \sdval{97.32\hp{0}}{0.06} & \sdval{0.82}{0.02} & \sdval{0.060}{0.001} & \sdval{177.0\hp{00}}{0.00}\\
            SRNODE\cpaper         & \sdval{100.00}{0.000} & \sdval{98.08\hp{0}}{0.15} & \sdval{1.24}{0.06} & \sdval{0.094}{0.003} & \sdval{259.0\hp{00}}{3.46}\\
            Local Unbiased ERNODE & \sdval{99.447}{0.039} & \sdval{97.526}{0.131} & \sdval{0.49}{0.002} & \sdval{0.046}{0.002} & \sdval{187.961}{1.812}\\
            Local Biased ERNODE   & \sdval{99.477}{0.051} & \sdval{97.488}{0.016} & \sdval{1.12}{0.065} & \sdval{0.044}{0.002} & \sdval{182.849}{1.578}\\
            \addlinespace
            \addlinespace
            Vanilla NSDE          & \sdval{98.27}{0.11} & \sdval{96.66}{0.16} & \sdval{2.70}{0.00} & \sdval{\hp{0}0.51}{0.07} & \sdval{313.86}{2.94}\\
            ERNSDE\cpaper         & \sdval{98.16}{0.11} & \sdval{96.27}{0.35} & \sdval{4.19}{0.04} & \sdval{\hp{0}7.23}{0.14} & \sdval{184.67}{2.31}\\
            SRNSDE\cpaper         & \sdval{98.79}{0.12} & \sdval{96.80}{0.07} & \sdval{8.54}{0.37} & \sdval{14.50}{0.40} & \sdval{382.00}{4.00}\\
            Local Unbiased ERNSDE & \sdval{98.05}{0.09} & \sdval{96.57}{0.13} &  \sdval{2.10}{0.01} & \sdval{\hp{0}0.39}{0.10} & \sdval{228.93}{1.77}\\
            Local Biased ERNSDE   & \sdval{98.02}{0.07} & \sdval{96.44}{0.16} & \sdval{1.90}{0.00} & \sdval{\hp{0}0.36}{0.03} & \sdval{230.10}{0.71}\\
            \bottomrule
        \end{tabular}
    }
    \caption{\textbf{MNIST Image Classification using Neural DE}: Using local unbiased regularization on neural ODE speeds up training by $\mathit{1.1\times}$ and predictions by $\mathit{1.9\times}$ while reducing the total NFEs to $\mathit{0.619\times}$. Local Biased Regularization tends to slow down training for smaller models on GPU while it further reduces the NFEs by $\mathit{0.602\times}$. For Neural SDE, we observe a similar reduction of NFEs by \timeschange{0.729}{0.733} and a training time improvement of \timeschange{1.28}{1.42}. The best global regularization method gets lower NFEs but overall takes more wall clock than the best performing local regularization method.}
    \label{tab:mnist_node}
    \vspace{-1em}
\end{table*}

\section{Experiments}
\label{sec:experiments}

In this section, we compare the effectiveness of unbiased and biased local regularization's effectiveness on the training and prediction timings of NDEs. We choose image classification and time series prediction problems in line with prior works on accelerating NDEs. We consider the following baselines:
\begin{enumerate}
    \item Vanilla Neural ODE with Continuous Interpolating Adjoint.

    \item Vanilla Neural SDE with discrete sensitivities.

    \item Global Regularization of Neural Differential Equations using discrete sensitivity analysis \citep{pal2021opening}.

    \item TayNODE~\citep{kelly2020learning} and STEER~\citep{ghosh2020steer} for models reported in \citet{pal2021opening}.
\end{enumerate}
\begin{figure}[t]
    \centering
    \includegraphics[width=0.8\linewidth]{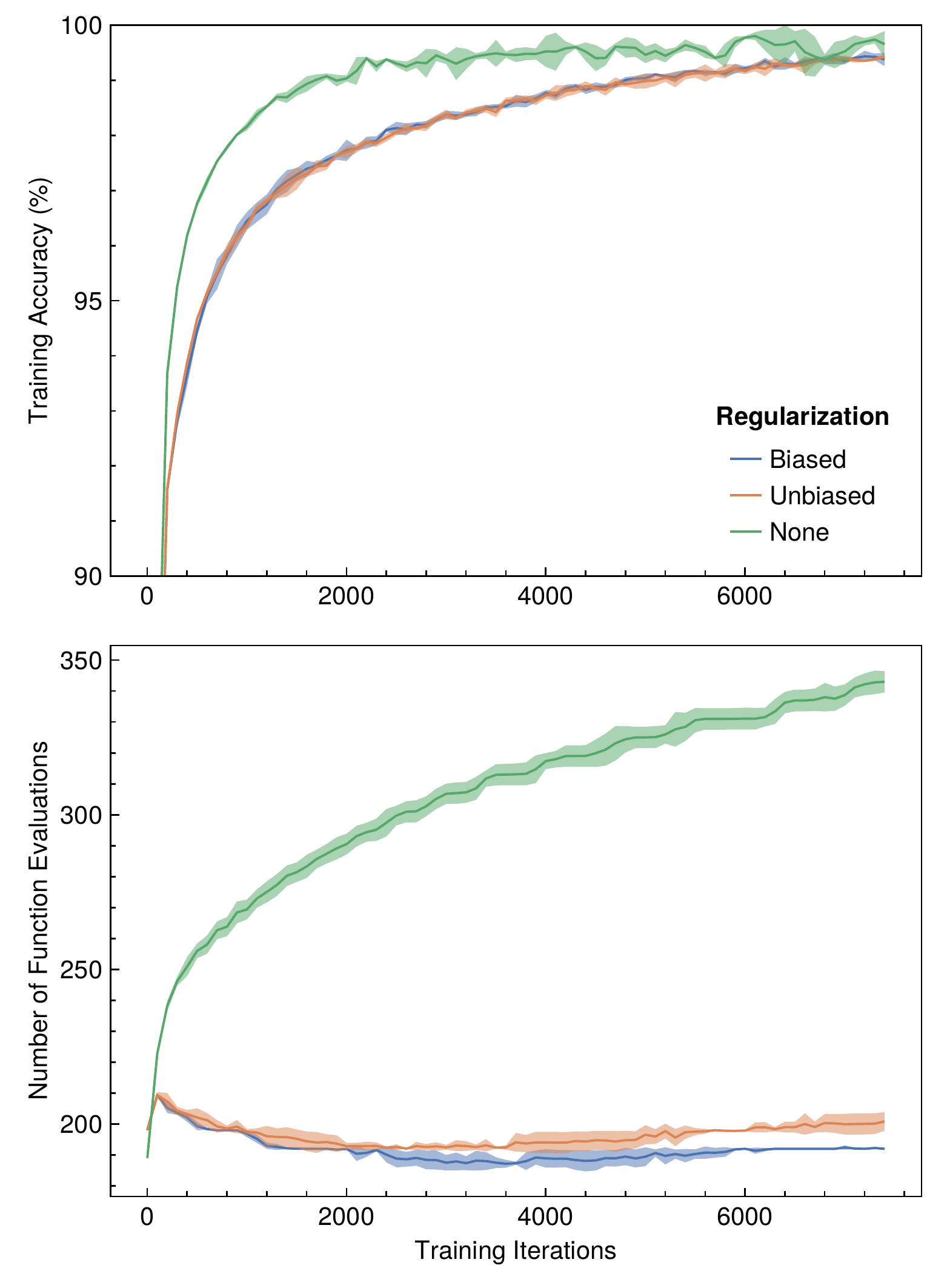}
    \caption{\textbf{MNIST Classification using Neural ODE}}
    \label{fig:mnist_node}
\end{figure}

We use the DifferentialEquations.jl~\citep{rackauckas2019diffeqflux} and Lux.jl~\citep{pal2022lux} software stack written in the Julia Programming Language~\citep{Julia-2017} for all our experiments.

Some details about the data presented in the tables:
\begin{itemize}
    \item All experimental results in the tables marked with \textparagraph~ were taken directly from \citet{pal2021opening}.

    \item We have tried to match the hardware details presented in the paper and the corresponding GitHub repository for \citet{pal2021opening}, but we note that differences in wall clock timings can be partially attributed to hardware.

    \item TayNODE~\citep{kelly2020learning} uses a different ODE integrator. Hence the NFEs are not directly comparable.
\end{itemize}

\begin{table*}[t]
    \centering
    \adjustbox{max width=0.9\textwidth}{
        \centering
        \begin{tabular}{lllll}
            \toprule
            \thead{Method} & \thead{Test Loss\\ ($\times 10^{-3}$)} & \thead{Training Time (hr)} & \thead{Prediction Time\\ (s / batch)} & \thead{Testing NFE}\\
            \midrule
            Vanilla NODE          & \sdval{3.41}{0.10} & \sdval{2.48}{0.22} & \sdval{0.16}{0.01} & \sdval{758.0}{25.87}\\
            STEER\cpaper          & \sdval{3.48}{0.01} & \sdval{1.62}{0.26} & \sdval{0.54}{0.06} & \sdval{699.0}{141.1}\\
            TayNODE\cpaper        & \sdval{4.21}{0.01} & \sdval{12.3}{0.32} & \sdval{0.22}{0.02} & \sdval{167.3}{11.93}\\
            ERNODE\cpaper         & \sdval{3.57}{0.00} & \sdval{0.94}{0.13} & \sdval{0.21}{0.02} & \sdval{287.0}{17.32}\\
            SRNODE\cpaper         & \sdval{3.58}{0.05} & \sdval{0.87}{0.09} & \sdval{0.20}{0.01} & \sdval{273.0}{0.000}\\
            Local Unbiased ERNODE & \sdval{3.64}{0.07} &  \sdval{2.31}{0.02} & \sdval{0.09}{0.00} & \sdval{422.0}{4.580}\\
            Local Biased ERNODE   & \sdval{3.63}{0.08} & \sdval{2.12}{0.24} & \sdval{0.10}{0.01} & \sdval{463.0}{63.02}\\
            \bottomrule
        \end{tabular}
    }
    \caption{\textbf{Physionet Time Series Interpolation}: Local Regularization reduces NFEs by \timeschange{0.556}{0.610} reducing the prediction timings by \timeschange{1.6}{1.78}. Our methods additionally improve training timings by \timeschange{1.073}{1.167}. We note that the difference in training time compared to (E/S)RNODE methods is due to change in the sensitivity algorithm.}
    \label{tab:physionet_node}
\end{table*}

\subsection{MNIST Image Classification}
\label{subsec:mnist}

We train a neural differential equation classifier to map flattened MNIST~\citep{lecun1998gradient} images to their corresponding labels.

\subsubsection{Neural Ordinary Differential Equation}
\label{subsubsec:mnist_node}

\textbf{Training Details:} We use the same model architecture as described in \citet{kelly2020learning}. Our model comprises of single hidden layered explicit model $f_\theta$ modeling the ODE dynamics followed by a linear classifier $g_\phi$.
The hidden layer is 100-dimensional. We train with a batch size of 512 for a total of 7500 steps. We use Adam~\citep{kingma2017adam} with a constant learning rate of $0.001$. For error estimate regularization, we exponentially decrease the regularization coefficient from $2.5$ to $1.0$. We use Tsit5~\citep{Tsit5} as the ODE integrator with an absolute and relative tolerance of $10^{-8}$.\footnote{We note that this is not a realistic tolerance at which image classification models are trained. We use this tolerance to allow a direct comparison to prior works.}

\textbf{Baselines:} We consider a Vanilla Neural ODE trained with the exact aforementioned specifications. All other baselines are directly taken from \citet{pal2021opening}.

\textbf{Results:} We summarize the results in \Cref{tab:mnist_node} and \Cref{fig:mnist_node}. Using local regularization speeds up prediction in all cases while it leads to a minor slowdown during training for biased sampling.

\vspace{-1.1em}

\subsubsection{Neural Stochastic Differential Equation}
\label{subsubsec:mnist_nsde}

\begin{figure}[t]
    \centering
    \includegraphics[width=0.8\linewidth]{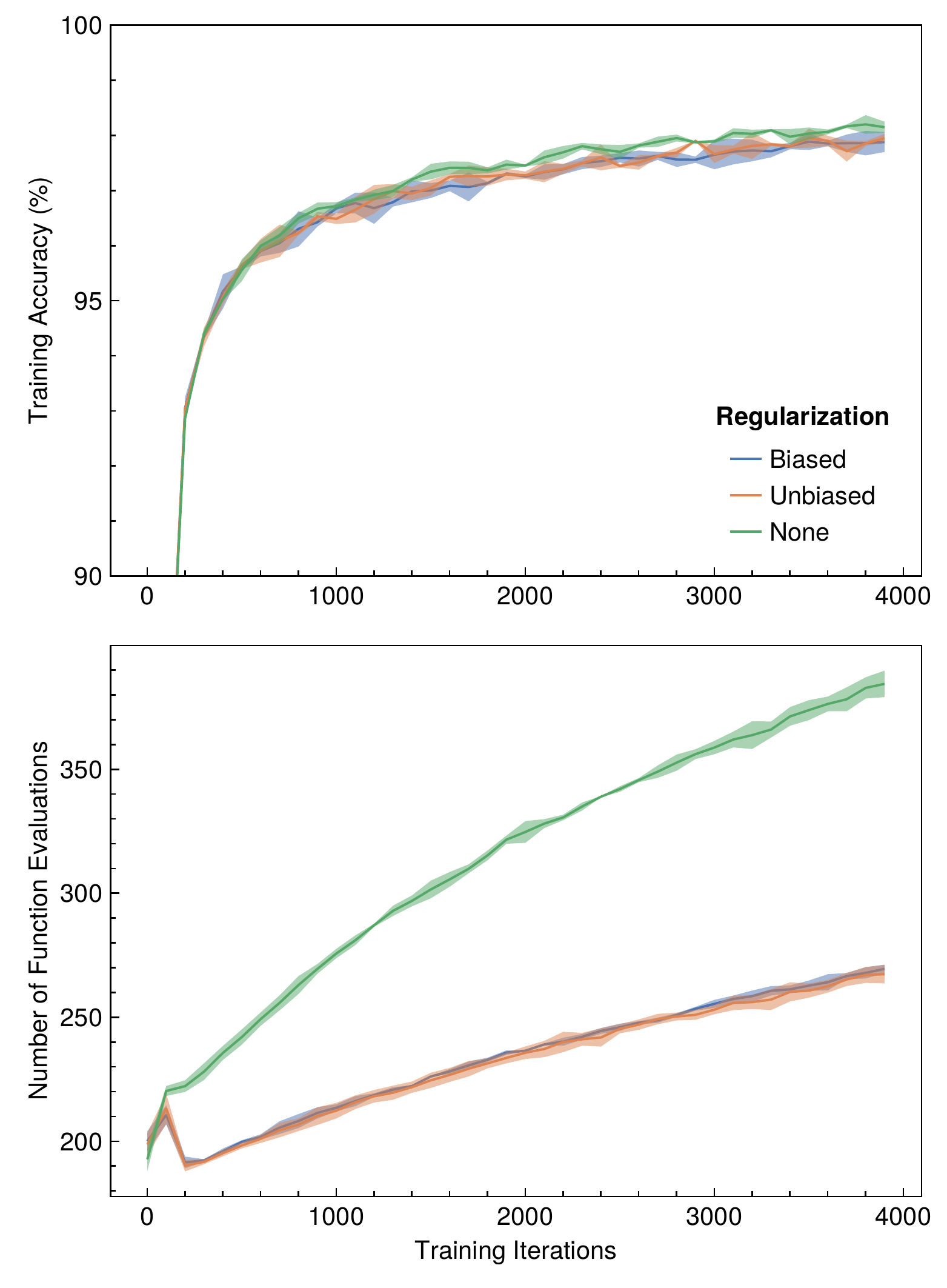}
    \caption{\textbf{MNIST Classification using Neural SDE}}
    \label{fig:mnist_nsde}
\end{figure}

\textbf{Training Details:} We downsample the flattened images to a 32-dimensional vector before feeding it into the Neural SDE which uses a diffusion model ($f_\theta$) having a 64-dimensional hidden layer and a linear drift model ($g_\phi$). Finally, a linear classifier ($h_\gamma$) predicts the label.
We train our models on CPU with a batch size of $512$ for a total of $4000$ steps. We optimize the weights using Adam~\citep{kingma2017adam} with a constant learning rate of $0.01$. We use SOSRI2 SDE solver~\citep{rackauckas2017adaptive} with a tolerance of $0.14$. We fix our regularization coefficient to be $10^3$. For this experiment, we rely on discrete sensitivity analysis.

\textbf{Baselines:} ERNSDE and SRNSDE results were taken from \citet{pal2021opening}. These were trained for $40$ epochs, nearly equivalent to training for $4000$ iterations. 

\textbf{Results:} We summarize the results in \Cref{tab:mnist_node} and \Cref{fig:mnist_nsde}. Local regularization improves training and prediction performance while keeping the test accuracy nearly constant.

\subsection{Physionet Time Series Interpolation}
\label{subsec:physionet}

\begin{figure}[t]
    \centering
    \includegraphics[width=0.8\linewidth]{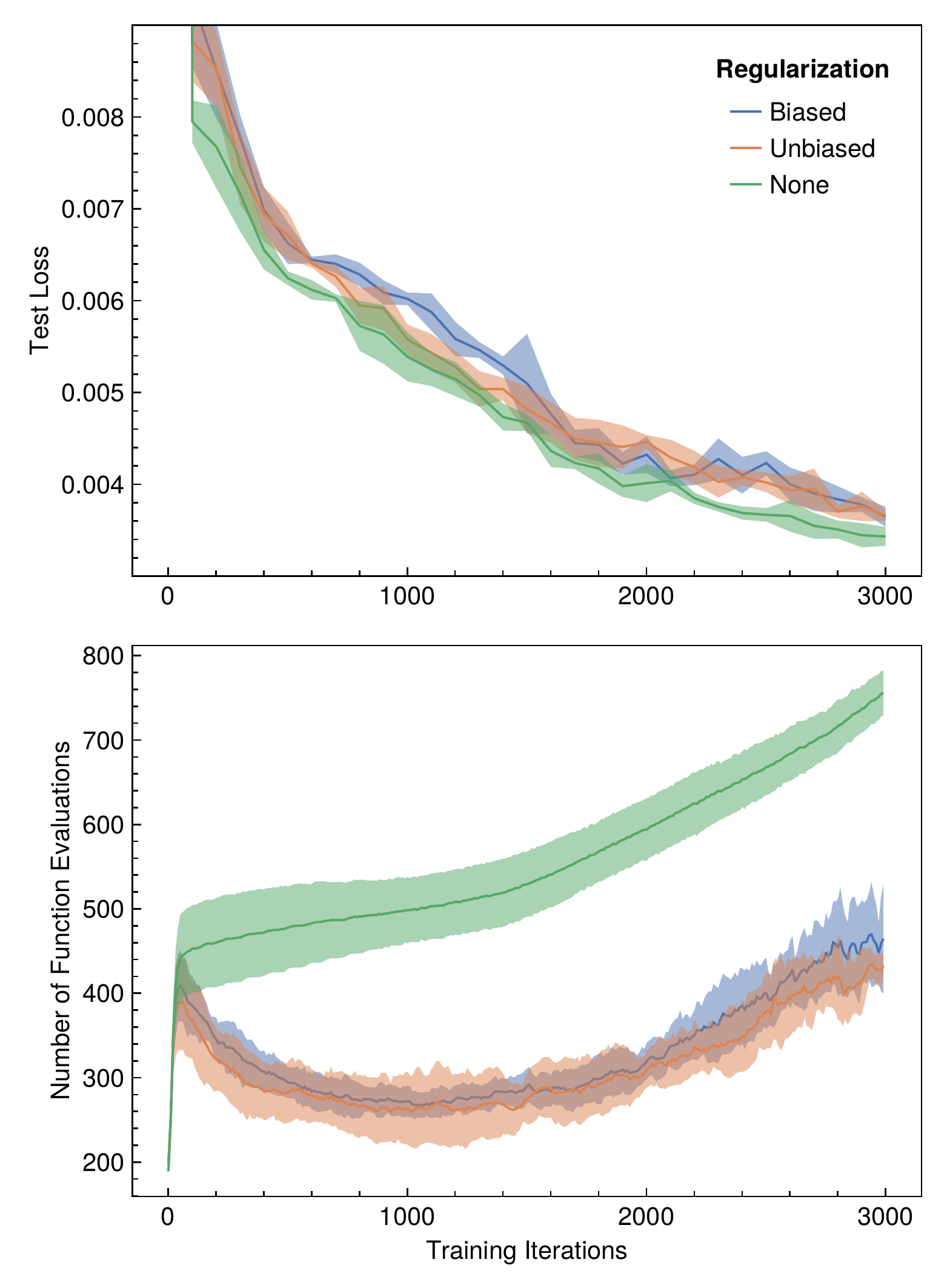}
    \caption{\textbf{Physionet Time Series Interpolation using Latent ODE}}
    \label{fig:physionet}
\end{figure}

\textbf{Training Details:} We use the experimental setup for Physionet 2012 Challenge Dataset~\citep{citi2012physionet} from \citet{kelly2020learning}. We use a Latent Neural ODE~\citep{rubanova2019latent} to perform time series interpolation on the dataset. We use the preprocessed dataset from \citet{kelly2020learning} to ensure a fair comparison and independent runs are performed using an 80:20 split of the dataset.

\begin{table*}[t]
    \centering
    \adjustbox{max width=0.9\textwidth}{
        \centering
        \begin{tabular}{lllllll}
            \toprule
            \thead{Configuration} & \thead{Method} & \thead{Train\\ Accuracy (\%)} & \thead{Test\\ Accuracy (\%)} & \thead{Training Time\\ (s / batch)} & \thead{Prediction Time\\ (s / batch)} & \thead{Testing\\ NFE}\\
            \midrule
            Standard & Vanilla           & \sdval{83.683}{1.450} & \sdval{67.394}{0.849} & \sdval{0.457}{0.018} & \sdval{0.130}{0.013} & \sdval{115.315}{12.136}\\
                     & Local Unbiased ER & \sdval{83.665}{0.805} & \sdval{67.678}{0.874} & \sdval{0.399}{0.014} & \sdval{0.096}{0.007} & \sdval{\hp{0}89.048}{\hp{0}7.335}\\
                     & Local Biased ER   & \sdval{83.958}{1.032} & \sdval{67.745}{0.824} & \sdval{0.555}{0.008} & \sdval{0.088}{0.003} & \sdval{\hp{0}81.301}{\hp{0}1.255}\\
            \addlinespace
            \addlinespace
            Multi-Scale & Vanilla           & \sdval{92.807}{12.458} & \sdval{80.048}{6.740} & \sdval{0.572}{0.012} & \sdval{0.170}{0.005} & \sdval{\hp{0}27.616}{\hp{0}0.905}\\
                       & Local Unbiased ER & \sdval{94.159}{\hp{0}9.694} & \sdval{80.432}{5.548} & \sdval{0.641}{0.025} & \sdval{0.175}{0.019} & \sdval{\hp{0}27.760}{\hp{0}0.177}\\
                       & Local Biased ER   & \sdval{99.987}{\hp{0}0.023} & \sdval{83.460}{0.727} & \sdval{0.774}{0.293} & \sdval{0.163}{0.015} & \sdval{\hp{0}26.334}{\hp{0}0.992}\\
            \bottomrule
        \end{tabular}
    }
    \caption{\textbf{CIFAR10 Image Classification using Neural DE}: For the standard Neural ODE, local regularization reduces the NFE by \timeschange{0.705}{0.772}, thereby improving prediction timings by \timeschange{1.35}{1.477}. However, unregularized model training takes $\mathit{0.823\times}$ the time for the biased model. For multi-scale models, the NFE and prediction time improvements are marginal and come at the cost of higher training time.}
    \label{tab:cifar10_node}
\end{table*}

For specific model architecture details, we refer the readers to \citet{pal2021opening}. We train the model for a total of $3000$ iterations using Adamax~\citep{kingma2017adam} with a learning rate of $0.01$ with $10^{-5}$ inverse decay per step. We use a batch size of $512$. We diverge from \citet{pal2021opening}, in using the regularization term as $\left(\eest\right)_{\treg} \cdot |dt|_{\treg}$ instead of the squared regularization term $\sum_j \left(\eest\right)_j^2$. Additionally, we decay the regularization coefficient exponentially from $100$ to $10$ over the $3000$ training iterations.

\textbf{Baselines:} Vanilla NODE was trained with the exact aforementioned configuration. All the other baselines were trained using discrete sensitivity analysis, and the exact details are present in \citet{pal2021opening}.

\textbf{Results:} We summarize the results in \Cref{fig:physionet} and \Cref{tab:physionet_node}.

\subsection{CIFAR10 Image Classification}
\label{subsec:cifar10}

\subsubsection{Neural Ordinary Differential Equation}
\label{subsubsec:cifar10_node}

\begin{figure*}[t]
    \centering
    \includegraphics[width=0.8\linewidth]{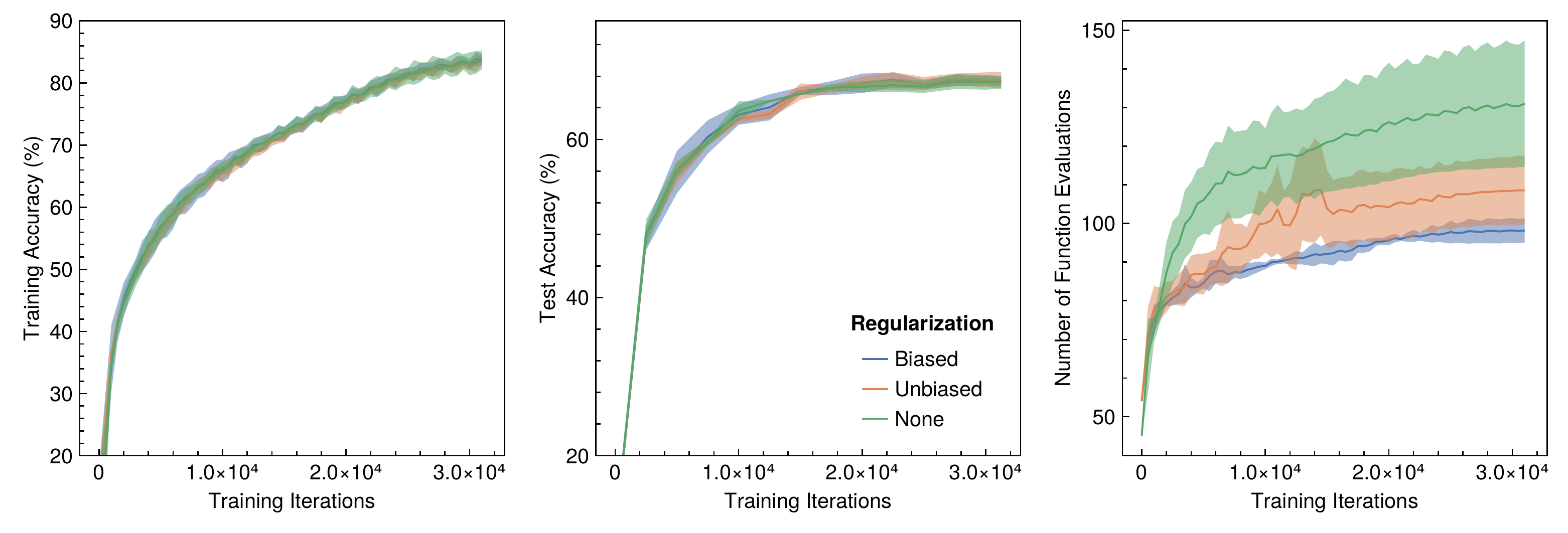}
    \caption{\textbf{CIFAR10 Image Classification using Standard Neural ODE}}
    \label{fig:cifar10_tiny}
\end{figure*}

\textbf{Training Details:} We use the CNN architecture for CIFAR10 as described in \citet{poli2020hypersolvers}. We train the models for $31250$ steps with Adam~\citep{kingma2017adam} using a cosine-annealing learning rate scheduler from $0.003$ to $0.0001$. We train the models with a batch size of $32$ and keep the regularization coefficient fixed at $2.5$. We use Tsit5~\citep{Tsit5} with a tolerance of $10^{-4}$.

\textbf{Results:} We summarize the results in \Cref{fig:cifar10_tiny} and \Cref{tab:cifar10_node}.

\subsubsection{Multiscale Neural ODE}
\label{subsubsec:cifar10_msnode}

\begin{figure*}[t]
    \centering
    \includegraphics[width=0.8\linewidth]{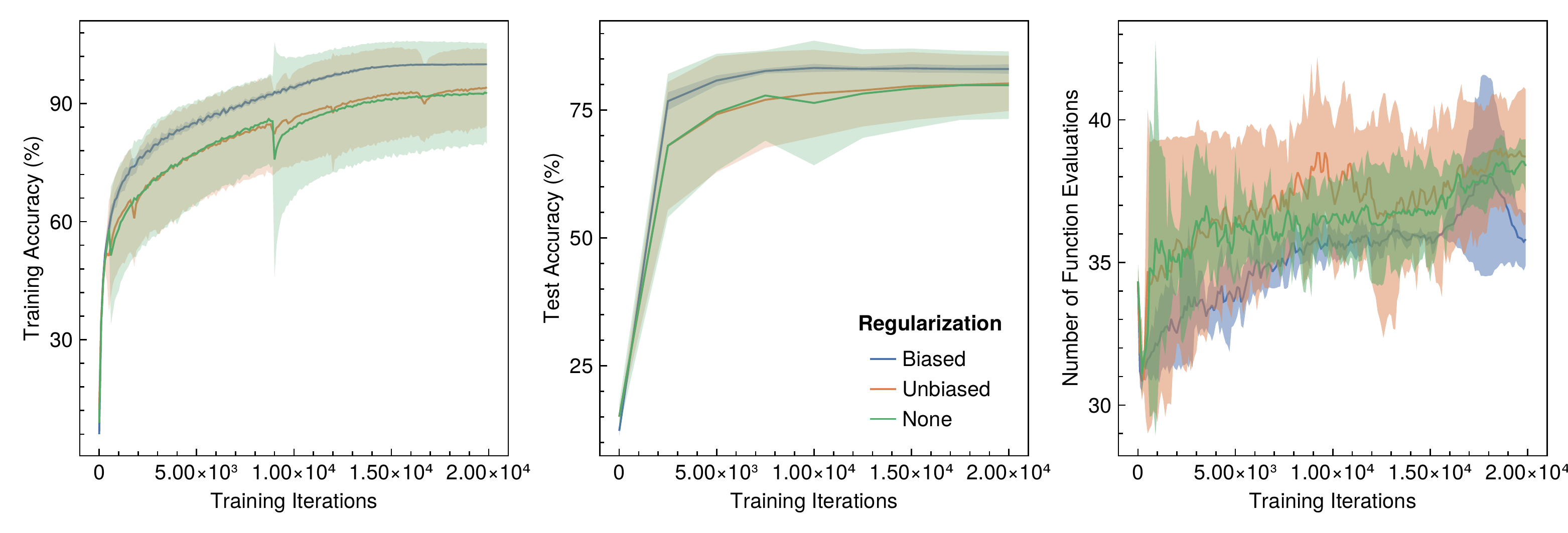}
    \caption{\textbf{CIFAR10 Image Classification using Multi-Scale Neural ODE}}
    \label{fig:cifar10_tiny_msnode}
\end{figure*}

\textbf{Training Details:} We modify the Tiny Multiscale DEQ architecture for CIFAR10 from \citet{bai_multiscale_2020} as Multiscale Neural ODE with Input Injection (See \Cref{subsec:multiscale_neural_ode}). To stabilize the training for larger models, we exponentially increase the regularization coefficient from $0.1$ to $5.0$. We train with a batch size of $128$ using VCAB3~\citep{wanner1996solving} with a tolerance of $0.05$.

\textbf{Results:} We summarize the results in \Cref{fig:cifar10_tiny_msnode} and \Cref{tab:cifar10_node}. The benefits from regularization for NFEs and prediction timings seem marginal. However, regularization using biased sampling makes the training dynamics stable as observed in \Cref{fig:cifar10_tiny_msnode}.

\section{Related Works}
\label{sec:related_works}

Neural Differential Equations (and by extension most Implicit Neural Networks) are mostly constrained by expensive training and prediction timings~\citep{dupont2019augmented, kelly2020learning, finlay2020train, ghosh2020steer, pal2021opening}. Traditionally accelerating these models have relied on using higher-order regularization terms to constrain the space of learnable dynamics~\citep{finlay2020train, kelly2020learning}. These models speed up predictions, but their benefits are often overshadowed by a massive training slowdown~\citep{pal2021opening}.

More recently, quite a few first-order schemes have been proposed. \citet{ghosh2020steer} randomized the endpoint of Neural ODEs to incentivize simpler dynamics. However, \citet{pal2021opening} didn't find significant benefits of using STEER in their experiments. \citet{pal2021opening} used internal solver heuristics -- local error and stiffness estimates -- to control the learned dynamics in a way that decreased both prediction and training time. We discuss their work in detail in \Cref{subsec:global_regularization}. \citet{xia2021heavy} rewrite Neural ODEs as heavy ball ODEs to accelerate both forward and backward passes. \citet{djeumou2022taylorlagrange} replace ODE solvers in the forward with a Taylor-Lagrange expansion and report significantly better training and prediction times.

\section{Discussion}
\label{sec:discussion}

In this manuscript, we have shown that we can obtain similar properties to global regularization by regularizing dynamical systems at randomly sampled time points. Additionally, this comes with the benefit of not being forced into a specific sensitivity analysis method. We have taken every experiment in \citet{pal2021opening} and empirically showed that our local regularization works at par with global regularization. However, our experiments using stiffness estimate for local regularization did not yield positive results and were not presented in this manuscript. Thus, we have demonstrated that we can ``close the blackbox'' and still leverage all the benefits of internal solver heuristics to improve training and predictions of neural differential equations.

\subsection{Limitations}
\label{sec:limitations}

We note the following limitations of our work:
\begin{itemize}
    \item Similar to \citet{pal2021opening}, if the objective is to learn the actual dynamical system, our method will not yield proper results. Our method is applicable only when the final state is relevant, i.e., in most classical deep learning tasks.

    \item Regularization introduces a new regularization coefficient hyperparameter which, if not tuned correctly, can lead to unstable dynamics or might negate the scheme's usefulness.
\end{itemize}

\section{Acknowledgement}
\label{sec:acknowledgement}

The authors acknowledge the MIT SuperCloud and Lincoln Laboratory Supercomputing Center for providing HPC resources that have contributed to the research results reported within this paper. This material is based upon work supported by the National Science Foundation under grant no. OAC-1835443, grant no. SII-2029670, grant no. ECCS-2029670, grant no. OAC-2103804, and grant no. PHY-2021825. We also gratefully acknowledge the U.S. Agency for International Development through Penn State for grant no. S002283-USAID. The information, data, or work presented herein was funded in part by the Advanced Research Projects Agency-Energy (ARPA-E), U.S. Department of Energy, under Award Number DE-AR0001211 and DE-AR0001222. We also gratefully acknowledge the U.S. Agency for International Development through Penn State for grant no. S002283-USAID. The views and opinions of authors expressed herein do not necessarily state or reflect those of the United States Government or any agency thereof. This material was supported by The Research Council of Norway and Equinor ASA through Research Council project ``308817 - Digital wells for optimal production and drainage''. Research was sponsored by the United States Air Force Research Laboratory and the United States Air Force Artificial Intelligence Accelerator and was accomplished under Cooperative Agreement Number FA8750-19-2-1000. The views and conclusions contained in this document are those of the authors and should not be interpreted as representing the official policies, either expressed or implied, of the United States Air Force or the U.S. Government. The U.S. Government is authorized to reproduce and distribute reprints for Government purposes notwithstanding any copyright notation herein.

\bibliography{main}
\bibliographystyle{icml2023}

\newpage
\appendix

\onecolumn

\end{document}